\documentclass[a4paper,conference]{IEEEtran}
\usepackage{amsmath}
\usepackage{amsfonts}
\usepackage{amssymb}
\usepackage{comment}
\usepackage{float}
\usepackage{caption}
\usepackage{color}
\usepackage{appendix}
\usepackage{hyperref}
\usepackage{graphicx}

\usepackage[caption=false,font=footnotesize]{subfig}
\usepackage{array}
\usepackage{cite}

\usepackage{xcolor}
\usepackage{soul}
\usepackage{multirow}

\begin{document}
\title{Image-based material analysis of\\ ancient historical documents}

\author{\IEEEauthorblockN{Thomas Reynolds}
\IEEEauthorblockA{Department of Computer Science\\
Royal Holloway, University of London\\
Email: thomas.reynolds.2021@live.rhul.ac.uk}
\and
\IEEEauthorblockN{Maruf A. Dhali}
\IEEEauthorblockA{Department of Artificial Intelligence\\
University of Groningen\\
Email: m.a.dhali@rug.nl}
\and
\IEEEauthorblockN{Lambert Schomaker}
\IEEEauthorblockA{Department of Artificial Intelligence\\
University of Groningen\\
Email: l.r.b.schomaker@rug.nl}}

\maketitle

\begin{abstract}
Researchers continually perform corroborative tests to classify ancient historical documents based on the physical materials of their writing surfaces. However, these tests, often performed on-site, requires actual access to the manuscript objects. The procedures involve a considerable amount of time and cost, and can damage the manuscripts. Developing a technique to classify such documents using only digital images can be very useful and efficient. In order to tackle this problem, this study uses images of a famous historical collection, the Dead Sea Scrolls, to propose a novel method to classify the materials of the manuscripts. The proposed classifier uses the two-dimensional Fourier Transform to identify patterns within the manuscript surfaces. Combining a binary classification system employing the transform with a majority voting process is shown to be effective for this classification task. This pilot study shows a successful classification percentage of up to 97\% for a confined amount of manuscripts produced from either parchment or papyrus material. Feature vectors based on Fourier-space grid representation outperformed a concentric Fourier-space format.
\end{abstract}


\section{Introduction}
\begin{figure}[!b]
\centering
\includegraphics[width=0.4\textwidth]{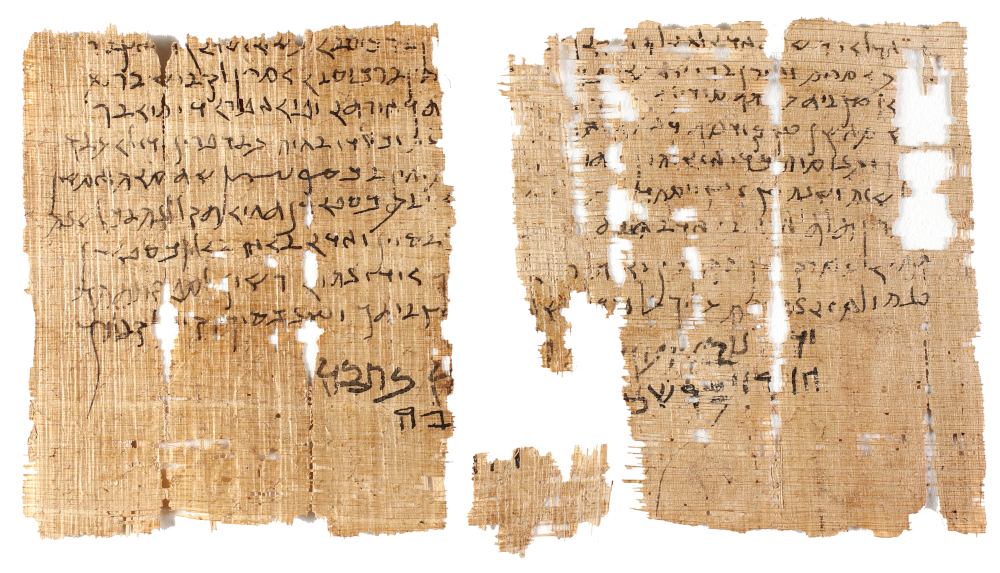}
\caption{Color image of plate X102 of the Dead Sea Scrolls collection containing three papyrus fragments. Distinctive striations can be seen in both the vertical and horizontal orientations. Damage is evident on the edges and within each fragment.}
\label{fig:x102}
\end{figure}

Image-based material classification is challenging due to large inter-class and intra-class variations within materials \cite{Kalliatakis2017}. Framing this problem in the context of ancient historical manuscripts provides a more significant challenge, primarily due to the degree of degradation of the data set. Gaining first-hand access to such manuscripts is often restricted or impractical. Subsequent chemical analysis of the material can also be damaging \cite{Freedman2018DestructiveResearchers}. On the contrary, analysis of the material using photographic images of manuscript samples causes no such damage. Furthermore, such images are relatively easy to produce and are often released into the public domain, allowing easier access. Previous material classification work has focused on inter material and texture classification techniques using data sets from non-context `clean' images \cite{Matsuyama1983StructuralTransformation} and data sets from `wild', context-set real-world images \cite{Bell2015MaterialDatabase}. Other work has incorporated material, texture, and pattern recognition techniques in specific real-world intra-material classification \cite{Wu2018FourierClassification, Kliangsuwan2018FFTImages, Camargo2009ImagePlants}. There has, however, been little usage of surface material classification techniques set in the context of photographic images of ancient manuscripts. This study uses images from the Dead Sea Scrolls (DSS) collection as a data set to investigate a classification method for materials of the writing surfaces (see figure \ref{fig:x102} for an example image). After conducting some pilot experiments with deep learning (convolutional neural nets), in the case of texture classification, a dedicated shape feature may prove to be more appropriate and convenient, particularly when considering the limited size of the available training data set. The research presented here employs a method in which the regular underlying periodic patterns inherent in the writing surface are used to classify the material of the manuscripts. Different feature vectors are constructed to capture these patterns.

The feature vectors are compared to specify which can accurately classify the writing materials of the manuscript fragments. These feature vectors are built upon the discrete 2-dimensional Fourier Transform (2DFT). Feature vectors which include the use of the 2DFT from both a computer vision and texture analysis stand-point, have been shown to provide standalone and complementary results to spatially focused approaches \cite{Hu2019FourierAnalysis}. For example, the 2DFT distinguishes between material textures and objects in non-contextual \cite{Bharati2004ImageComparisons,Bajcsy1973ComputerSurfaces} and contextual images \cite{CEVIKALP2017TheClassification} in conjunction with a standalone classifier \cite{Hui2014DiscreteClassifiers} or input to a neural network \cite{Franzen2019VisualizingDomain, Kumar20212D-FFTNetwork}. Furthermore, incorporating the 2DFT into Compositional Pattern Producing Networks (CPPNs) has shown improvements in accurately recreating textures due to the ability of the 2DFT to match and capture high-frequency details \cite{Tesfaldet2019Fourier-CPPNsSynthesis}. In addition, the Fourier Transform provides a more straightforward approach than Markov Random Field modeling of textures \cite{Hassner1980TheTexture}. Thus, 2DFT feature vectors offer an attractive solution to distinguish between inherent textures and patterns in the writing surface of ancient historical manuscripts, such as the DSS, to classify the materials upon which they are written.

\subsection{Dead Sea Scrolls}
\label{sec:_dss}
The DSS collection consists of ancient historical manuscripts produced between the third century BCE and the first century CE, written mainly on parchment (animal skin) and papyrus (made from the pith of the papyrus plant) and, as a singular exception, copper \cite{Shor2014TheScrolls}. Recent works on the DSS have concerned handwriting analysis, dating of the scrolls and writer identification \cite{Dhali2017AScrolls, Dhali2019BiNet:Networks, Dhali2020Feature-extractionDevelopment, popovic2021artificial}. More recent work on the DSS related to the materials used in their production has focused on matching manuscript fragments. In this approach, a neural network is employed to fit manuscript fragment pairs, which potentially originate from the same sheet of papyrus \cite{Abitbol2021MachinePapyrus}. The approach utilizes patterns found in papyrus material and demonstrates some of the difficulties of working with damaged ancient historical manuscripts. In other work, the materials on which these manuscripts have been written, are analyzed by material scientists employing micro and macro x-ray fluorescence imaging, scanning electron microscopy, spectroscopy and microchemical testing \cite{WolffProvenanceMicro-XRF,Rabin2013Archaeometry,Loll2019MuseumAnalysis}. Answers to questions probing the provenance and archaeometry of the DSS are a result of such studies, and are based upon the correct identification of the underlying material. Utilizing such methods may not always be feasible due to cost, personnel availability, potential damage to the manuscripts, unavailability of technology and time. Instead, a pattern recognition system may help to classify the writing materials of the manuscripts while mitigating some of the traditional impracticalities, and by extension, help in the pursuit of answers to such questions. Despite discoloration and damage to the manuscripts over time which hinders accurate and expedient classification, the underlying periodic and regular patterns found in the material remain, and may form the basis of a classification system (see figure \ref{fig:x102}). Testing the accuracy of such a system can help determine whether traditional material analysis techniques used on the DSS and other ancient manuscripts have the potential for supplementation by such a system, or by a further extension of one.

This work mainly focuses on classifying the primary writing materials but opens the door for further in-depth analysis using pattern recognition techniques.

\section{Methodology}
This section will briefly present the data, preprocessing measures, sampling techniques, feature vector construction, and finally, the classification step. 

\subsection{Data}
\begin{figure}[!t]
\centering
\includegraphics[width=0.4\textwidth]{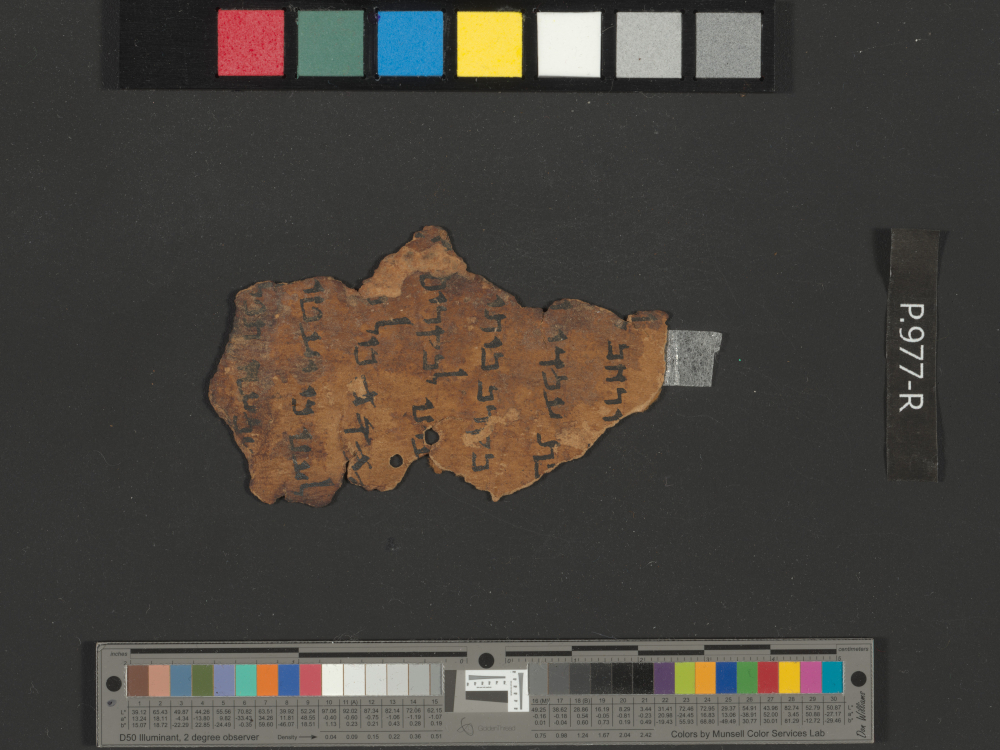}
\caption{A multi-spectral color image of plate 977 containing a single fragment made of parchment. Color calibration panels, scale bars, and plate label are visible in the image.}
\label{fig:977}
\end{figure}

The data set consists of DSS images kindly provided by the Israel Antiquities Authority (IAA). The images are publicly available on the website of the Leon Levy Dead Sea Scrolls Digital Library project \cite{Shor2014TheScrolls}. The DSS collection comprises primarily two types of material; mostly parchment with a minority written on papyrus, and a singular exception being a copper plate. This study excludes the copper plate and uses parchment and papyrus examples resulting in a binary classification task. The vast majority of the scrolls have experienced some form of degradation due to aging or mishandling. It is common for parts of a scroll to be missing, of which a few disconnected fragments remain for analysis. Some of the remaining fragments are of relatively small size and vary in condition. Some of the images are available in a range of spectral bands. This study uses two sets of scrolls images: ordinary color photographs (Set-1) and composite images composed of the individual spectral band images (Set-2). Figures \ref{fig:x102} and \ref{fig:977} show examples from Set-1 and Set-2 respectively. The sets are used independently. There are notable differences between the two sets of images. While Set-1 images vary in size and distance to subject, Set-2 images are $7216 \times 5412$ pixels and do not vary in distance. Some Set-1 images contain more than one fragment of scroll due to the nature of the arrangement by the curators. Some Set-2 MS images are only partial images of a larger fragment. The images used for this study are a subset of the complete collection. The subset was made based on two criteria;  to represent as many of the different textures available throughout the collection as possible and, secondly, to maximize the available area of material for samples to be taken. The images are comprised of 23 parchment fragments and 10 papyrus fragments, totaling 33 fragments in each set (table \ref{tab:samples}). The same fragments were used for both sets of images (For further details, please see the table in Supplementary Material A).

\begin{table}[!h]
\caption{Fragments Count by Material}
\centering
\begin{tabular}{ccc}
\hline \hline
Parchment & Papyrus & Total \\ \hline \hline
23        & 10      & 33    \\ \hline
\end{tabular}
\label{tab:samples}
\end{table}

\subsection{Preprocessing}
Identifying each fragment within an image is the first step in preprocessing. This is a difficult step, especially for the Set-2 images, containing color reference panels and measurement markings (figure \ref{fig:977}). The fragments in an image are identified from the background, other fragments, and the reference markings, using automated k-means clustering and by hand. Then, each fragment is extracted for individual processing.

\begin{figure}[!t]
\centering
\includegraphics[width=0.42\textwidth]{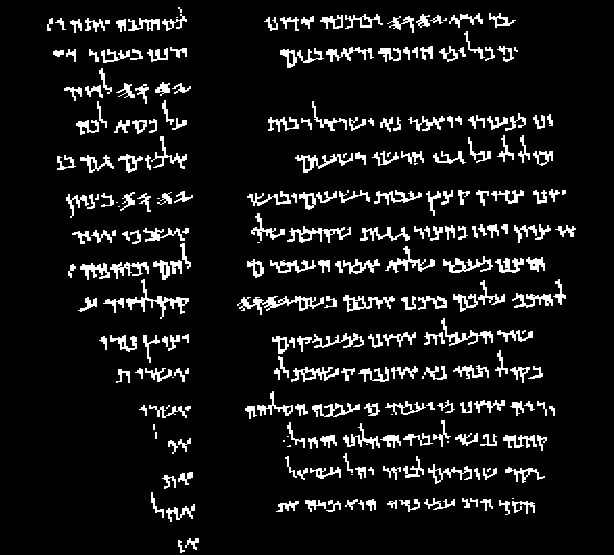}
\caption{A binarized image showing the ink (text) from plate 976. The binarization is obtained using the BiNet network \cite{Dhali2019BiNet:Networks}.}
\label{fig:undilated}
\end{figure}

In order to extract features, clean images of the writing surface material within each fragment are required. Removal of the text prevents any regular and periodic patterns within the text from influencing those patterns found in the material. Combining the need for clean background material with the limited supply of DSS material, the text and gaps caused by damage within the boundaries of each fragment are filled to provide more sample material. Previous work using deep learning \cite{Dhali2019BiNet:Networks} has produced binary masks of only the visible text in each image (figure \ref{fig:undilated}). These have kindly been provided for use with this study. As each binary text image overlays onto the original image, these masks identify text locations within each fragment. These masks are made robust by dilation; one pixel of text in the mask is expanded to a $3 \times 3$ pixel square (For further details, please see the figure in Supplementary Material B). This is a necessary alteration as the masks were used to analyze the written text of the scrolls, having been designed to capture no background material strictly.

\begin{figure}[!t]
\centering
\includegraphics[width=0.4\textwidth]{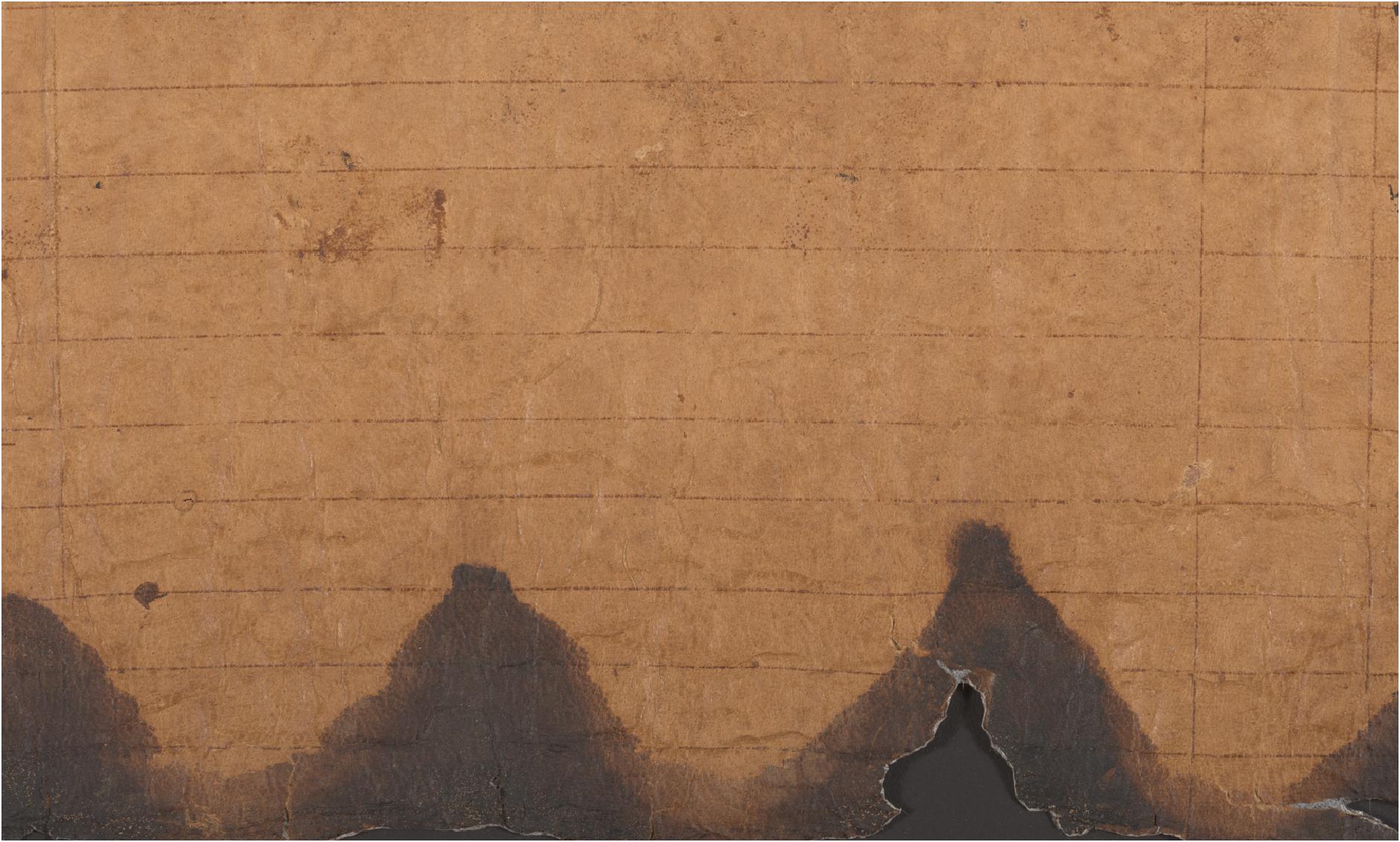}
\caption{A zoomed view \textit{after} the filling process; taken from the multi-spectral image of plate 974 (Set-2)}
\label{fig:fill}
\end{figure}

\begin{figure}[!t]
\centering
\includegraphics[width=0.38\textwidth]{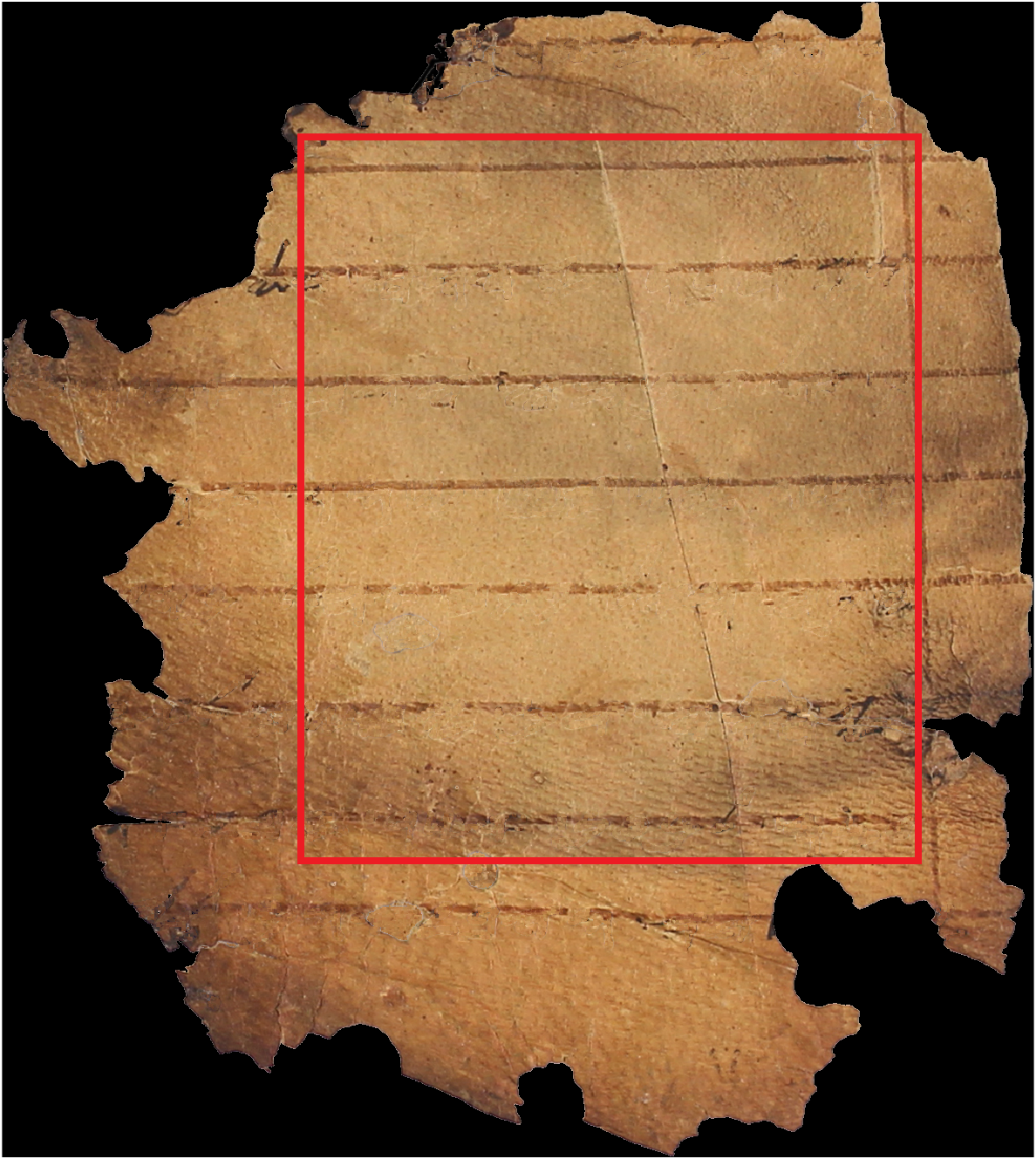}
\caption{Sample area from a fragment extracted from the color image of plate 1039-2 (Set-1). The largest internal inscribed rectangle (by area) is found. Samples used for feature vector creation will be taken from this area at evenly spaced intervals.}
\label{fig:inscribed}
\end{figure}

Thus, in a minority of instances, the outline of the written text may remain. Dilation of the text pixels helps capture any text not included in the original binary masks. Maintaining the regular patterns found in the material is of primary concern. The method chosen for filling in these locations was selected to maintain the surface patterns and is known as exemplar infilling \cite{Criminisi2004RegionInpainting}. Exemplar infilling searches the entirety of the fragment-image for a patch of material that matches a section of the location to be filled most closely, based on the sum of squared differences. This patch is then used to fill that section (figure \ref{fig:fill}). Each location is filled using different, closely matched patches of specified size ($9 \times 9$ pixels). A new patch is searched for after every iteration, as the closest matched patches may change once the filling process starts. In addition, most of the damage to each scroll fragment is found around the edges. In order to mitigate the influence of degradation on the classification step, samples are taken from the interior part of the fragment. This is achieved by considering only the portion contained within the largest inscribed rectangle (by area) within the fragment (figure \ref{fig:inscribed}). This area is known as the sample area. Samples for feature vector construction are taken from within this area.

\begin{figure}[!t]
\centering
\includegraphics[width=0.4\textwidth]{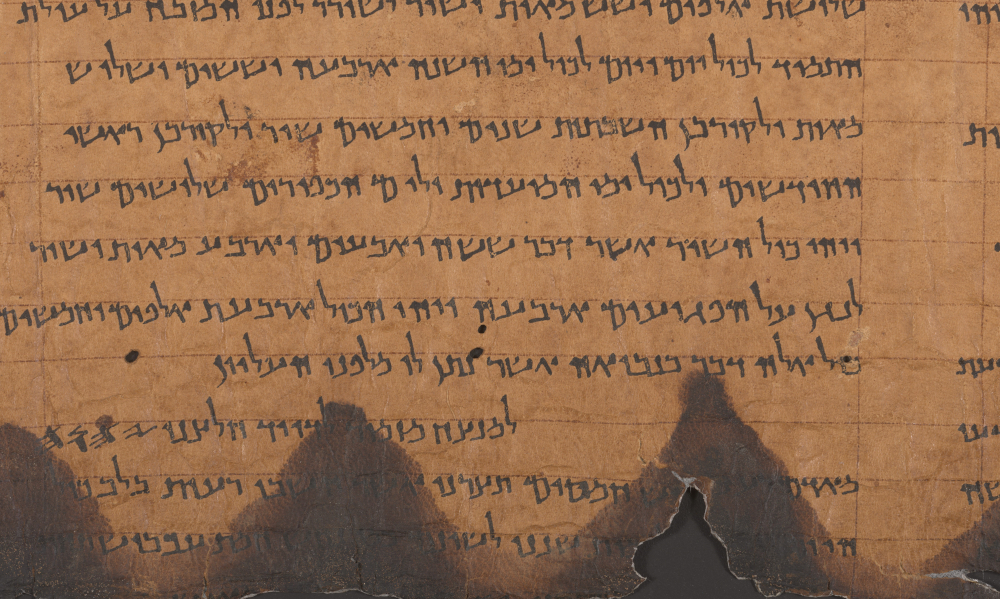}
\caption{A zoomed view \textit{before} the filling process; taken from the multi-spectral image of plate 974 (Set-2)}
\label{fig:prefill}
\end{figure}

\subsection{Sampling}
Twenty-five samples of size $256 \times 256$ pixels are taken at evenly spaced intervals in a $5 \times 5$ grid pattern covering the sample area. As each sample area differs in size, the spacing between samples depends on the area's dimensions. As a result, there is an overlap of the samples for smaller sample areas, and conversely, there is unused space between samples for large sample areas. This situation is unavoidable due to the inconsistent nature of the fragment sizes.

\subsection{Feature Vector Construction}
\begin{figure}[!t]
\centering
\includegraphics[width=0.4\textwidth]{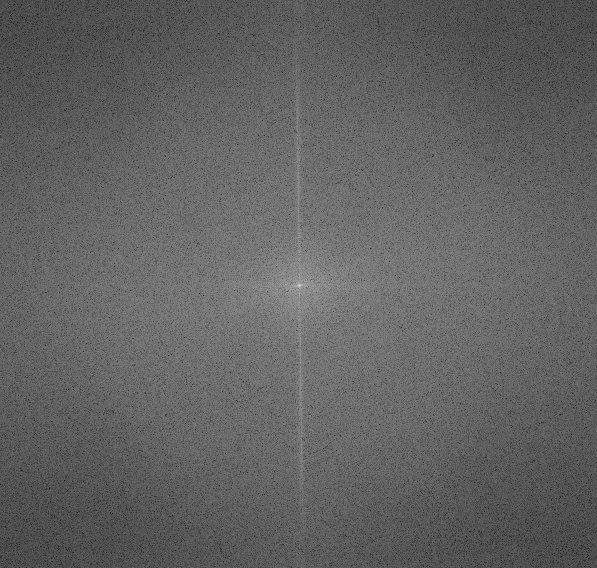}
\caption{Visual representation of the log2DFT applied to a parchment-image sample.}
\label{fig:ftlog}
\end{figure}

\begin{figure}[!t]
\centering
\includegraphics[width=0.4\textwidth]{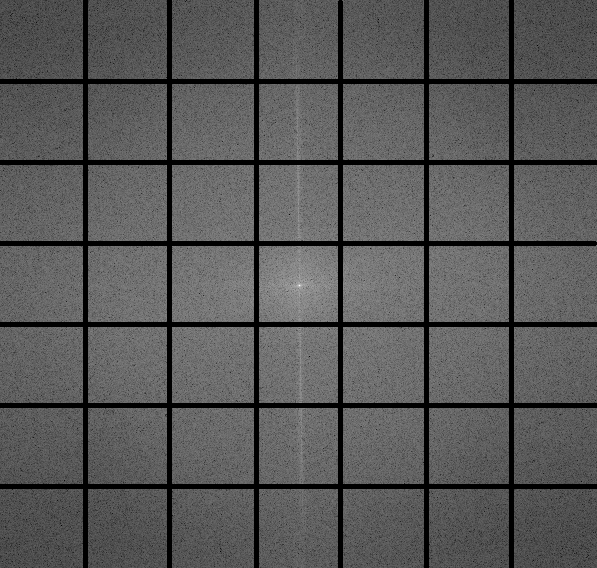}
\caption{The grid is used to create the primary feature vectors. The mean and standard deviation of the pixel values in each grid area are used as the basis for the feature vectors.}
\label{fig:ftgrid}
\end{figure}

\begin{figure}[!t]
\centering
\includegraphics[width=0.4\textwidth]{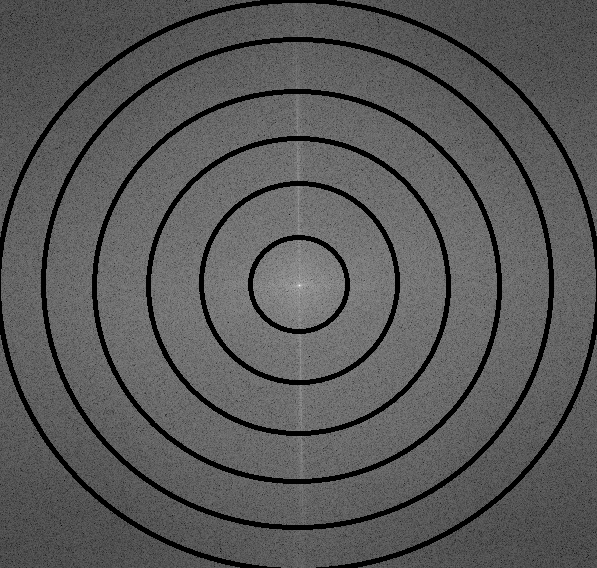}
\caption{The concentric ring division is used in construction of secondary feature vectors.  The mean and standard deviation of the pixel values in ring area are used as the basis for the feature vectors.}
\label{fig:circles}
\end{figure}

The saturation values of each sample image, from an RGB to HSV conversion of the image, are extracted. The extraction assigns each pixel a value from 0 to 1 to describe its saturation level. The use of HSV and saturation values have been shown to provide improved performance when using the 2DFT for image classification purposes \cite{Kliangsuwan2018FFTImages, Wu2018FourierClassification}. The 2DFT is then applied. The DC component is centralized, and the log transform of the absolute values is taken (log2DFT) to visually display the spectrum (figure \ref{fig:ftlog}). The log2DFT representation is then partitioned into $n \times n$ non-overlapping sections (figure \ref{fig:ftgrid}). This study uses $n = 7$. As there has been little previous work concerning the use of the log2DFT on ancient historical manuscripts, this value was a subjective decision made to balance the feature vector detail level with the feature vector length. The mean of the values in each section is calculated, and the values are concatenated to produce a feature vector for the sample image. This process is repeated for the standard deviation of the pixels, resulting in two separate feature vectors for a sample (mean feature vector as MFV and standard deviation feature vector as SDFV). These are known as the primary feature vectors. In addition, three secondary feature vectors are proposed, making five in total. The first is a feature vector based on dividing the log2DFT into six concentric rings (figure \ref{fig:circles}). Similar to the primary feature vectors, the mean and standard deviation of the pixel values in each ring are calculated and separately concatenated to produce two feature vectors. The decision to include a concentric ring feature vector is based on the rotational variance of the log2DFT, with the orientation of the sample potentially affecting classification performance. A final feature vector based on work by \cite{CEVIKALP2017TheClassification} is trialed. This method uses each pixel's magnitude and phase angle from a log2DFT. The magnitude acts as a weighted vote and deposits the pixel into one of nineteen phase angle bins, evenly spaced from $0$ to $2\pi$. The value across all bins is normalized to one and concatenated to produce a $1 \times 19$ feature vector for the sample.

\subsection{Classification}
Each of the five proposed feature vectors is handled independently in the same way. Considering only one feature vector at a time, each sample has its associated feature vector and ground truth, parchment or papyrus, stored in a dictionary. There are 825 individual dictionary entries. A leave-one-out method is then employed. All 25 feature vectors of a fragment are removed from the dictionary. These feature vectors are now effectively unseen. Each of the 25 removed vectors is compared against the remaining set stored in the dictionary to find its closest match. This match is based on the Euclidean distance between the two vectors. Once the closest feature vector in the dictionary has been found, its associated ground truth is recorded. The number of matches labeled as parchment is compared to the number labeled as papyrus, providing a percentage of belief as to the fragment's material. The fragment is classified based on the greater percentage. This is repeated for all fragments. The $F$-scores at both the fragment and sample levels are calculated. The traditional balanced $F_1$ score is used (eq. \ref{eq:F}).
\begin{equation}\label{eq:F} 
F_1 = 2\cdot\frac{precision\cdot recall}{precision + recall}
\end{equation}

\section{Results}
In total, 33 fragments (23 parchment, 10 papyrus) were used. Table \ref{tab:overall} shows the overall classification success percentage for the image types using the primary feature vectors MFV and SDFV. Tables \ref{tab:conf_mean} and \ref{tab:conf_sd} show the confusion matrices for the results of the primary feature vectors. Tables \ref{tab:ffrag} and \ref{tab:fsamp} show the precision, recall and $F$-score for classification at the fragment and sample levels for the primary feature vectors. The three secondary feature vectors showed less successful results (For further details, please see the tables in Supplementary Material B).

\begin{table}[!ht]
\caption{Classification Success (\%) for Primary Feature Vectors (MFV and SDFV)}
\centering
\begin{tabular}{lcc}
\hline \hline
\textbf{Image type} & \textbf{MFV}  & \textbf{SDFV} \\ \hline \hline
Color               & 90.9          & 90.9          \\ \hline
Multispectral       & \textbf{97.0} & \textbf{97.0} \\ \hline
\end{tabular}
\label{tab:overall}
\end{table}

\begin{table}[!ht]
\caption{Confusion Matrix (\%) for the MFV}
\centering
\begin{tabular}{llcc}
\hline \hline
\textbf{Image type }           & \textbf{True class} & \multicolumn{2}{c}{\textbf{Classified as}} \\ \hline \hline
                               &            & Parchment        & Papyrus        \\ \hline
\multirow{2}{*}{Color}         & Parchment  & \textbf{100.0}   & 0.0            \\ \cline{2-4} 
                               & Papyrus    & 30.0             & \textbf{70.0}  \\ \hline
\multirow{2}{*}{Multispectral} & Parchment  & \textbf{100.0}   & 0.0            \\ \cline{2-4} 
                               & Papyrus    & 10.0             & \textbf{90.0 } \\ \hline
\end{tabular}
\label{tab:conf_mean}
\end{table}

\begin{table}[!ht]
\caption{Confusion Matrix (\%) for the SDFV}
\centering
\begin{tabular}{llcc}
\hline \hline
\textbf{Image type}            & \textbf{True class} & \multicolumn{2}{c}{\textbf{Classified as}} \\ \hline \hline
                               &            & Parchment        & Papyrus        \\ \hline
\multirow{2}{*}{Color}         & Parchment  & \textbf{95.7}    & 4.3            \\ \cline{2-4} 
                               & Papyrus    & 20.0             & \textbf{80.0}  \\ \hline
\multirow{2}{*}{Multispectral} & Parchment  & \textbf{100.0}   & 0.0            \\ \cline{2-4} 
                               & Papyrus    & 10.0             & \textbf{90.0}  \\ \hline
\end{tabular}
\label{tab:conf_sd}
\end{table}

\begin{table}[!ht]
\caption{Precision, Recall and $F$-score for Classification at the Fragment Level}
\centering
\begin{tabular}{llccc}
\hline \hline
\textbf{Image type}            & \multicolumn{4}{l}{\textbf{Mean Feature Vector (MFV)}}                \\ \hline \hline
                               & Material       & Precision       & Recall          & F-score          \\ \hline
\multirow{2}{*}{Color}         & Parchment      & 0.88            & \textbf{1}      & 0.94             \\ \cline{2-5} 
                               & Papyrus        & \textbf{1}      & 0.70            & 0.82             \\ \hline
\multirow{2}{*}{Multispectral} & Parchment      & 0.96            & \textbf{1}      & \textbf{0.98}    \\ \cline{2-5} 
                               & Papyrus        & \textbf{1}      & 0.90            & 0.95             \\ \hline \hline
\textbf{Image type}            & \multicolumn{4}{c}{\textbf{Standard Deviation Feature Vector (SDFV)}} \\ \hline \hline
\multirow{2}{*}{Color}         & Parchment      & 0.92            & 0.96            & 0.94             \\ \cline{2-5} 
                               & Papyrus        & 0.89            & 0.80            & 0.84             \\ \hline
\multirow{2}{*}{Multispectral} & Parchment      & 0.96            & \textbf{1}      & \textbf{0.98}    \\ \cline{2-5} 
                               & Papyrus        & \textbf{1}      & 0.90            & 0.95             \\ \hline
\end{tabular}
\label{tab:ffrag}
\end{table}

\begin{table}[!ht]
\caption{Precision, Recall and $F$-score for Classification at the Sample Level}
\centering
\begin{tabular}{llccc}
\hline \hline
\textbf{Image type}            & \multicolumn{4}{l}{\textbf{Mean Feature Vector (MFV)}}                \\ \hline \hline
                               & Material     & Precision        & Recall           & F-score          \\ \hline
\multirow{2}{*}{Color}         & Parchment    & 0.85             & 0.91             & 0.88             \\ \cline{2-5} 
                               & Papyrus      & 0.77             & 0.64             & 0.70             \\ \hline
\multirow{2}{*}{Multispectral} & Parchment    & \textbf{0.93}    & \textbf{0.96}    & \textbf{0.95}    \\ \cline{2-5} 
                               & Papyrus      & 0.91             & 0.84             & 0.87             \\ \hline \hline
\textbf{Image type}            & \multicolumn{4}{c}{\textbf{Standard Deviation Feature Vector (SDFV)}} \\ \hline \hline
\multirow{2}{*}{Color}         & Parchment    & 0.85             & 0.89             & 0.87             \\ \cline{2-5} 
                               & Papyrus      & 0.72             & 0.64             & 0.68             \\ \hline
\multirow{2}{*}{Multispectral} & Parchment    & \textbf{0.91}    & \textbf{0.96}    & \textbf{0.93}    \\ \cline{2-5} 
                               & Papyrus      & 0.89             & 0.78             & 0.83             \\ \hline
\end{tabular}
\label{tab:fsamp}
\end{table}

\section{Discussion}
This study presented a binary 2DFT based technique to classify material used in ancient historical manuscripts, specifically the DSS. This technique, concerning the primary feature vectors (based on Fourier-space grid representation) used in conjunction with the multi-spectral images, showed a relatively high level of performance with regards to overall classification percentage ($\approx 97\%$ compared to $\approx 91\%$ successful classification for MS and color images respectively) across both primary feature vectors. In addition, the mean feature vector produced slightly improved results than the standard deviation feature vector. This was particularly true for classification accuracy as measured by $F$-score values at the sample level. The papyrus images were, in general, more susceptible to misclassification than parchment images, showing lower recall and $F$-scores in their classification than parchment samples (tables \ref{tab:ffrag} and \ref{tab:fsamp}). 

When combined with multi-spectral images and using the MFV, the proposed technique achieved the highest accuracy. This is also the case for color images. This result may prove helpful for future work involving manuscripts that have not been photographed using multi-spectral equipment. Several reasons may explain the difference in performance between the multi-spectral and the color images. For example, the multi-spectral images reveal more discriminatory patterns, periodic frequencies, magnitudes, and details in the materials of the image compared to the color images. This arises from the different wavelengths of light highlighting different material details. The recombination of the images of different wavelengths may have captured more features that make each material unique for the log2DFT method. The multi-spectral images were also of higher resolution, enabling the visual-based technique to discriminate more clearly between materials. By investigating the different spectral band images individually, the band that provides the best results could be used in further research. This approach has potential implications for photographing decisions concerning other manuscripts. The MFV performed at least as well as the SDFV regarding $F$-score. This may suggest that measures of the spread between pixel values in grid sections provide less discriminatory ability than measures concerning the value of the pixels. 

Further work may investigate which particular frequencies and associated magnitudes provide improved discrimination between particular materials. This study used a selection of images to create the dictionary of feature vectors. These images were chosen based on a subjective view that they captured a high proportion of the different types of textures of parchment and papyrus and represented the DSS collection as a whole. However, there were fewer papyrus examples in the set than parchment. As a result, the papyrus samples proved more challenging to classify accurately. By expanding the dictionary set, an improved representation of the collection can be achieved, which could yield higher classification results, particularly about improving the classification of papyrus fragments. Therefore, more examples would be encountered, and closer matches to novel samples may be made. The measured voting procedure would then use more votes to confirm a classification, impacting the belief percentage per sample. The use of binary text masks was beneficial to fill the images. However, access to such materials may not always be possible. In a small minority of cases, most notably affecting some papyrus manuscripts, some of the binary images did not capture all the text on the manuscripts. This could have influenced the results, mainly where we see a slightly worse performance for the classification of the papyrus fragments which exhibited more residue text post fill. A fixed number of samples and a fixed log2DFT feature vector grid were used in this study. A suggestion for further work would be to investigate whether changing these parameters could improve the results. The secondary, i.e., the concentric Fourier-space feature vector results show relatively poorer performance than the primary feature vectors, particularly concerning the papyrus fragments (For further details, please see Supplementary Material C). With reference to the modified weighted bin feature vector, which showed success in the study by \cite{CEVIKALP2017TheClassification}, this was particularly evident, achieving a low $10\%$ success rate for papyrus images (For further details, please see the tables in Supplementary Material B). 

The image data set used in the research consisted of real-world objects set in a context, which may not have been appropriate for this study. This study's relatively simplistic periodic patterned images may not have provided enough discriminatory differences for this method to work effectively. The alternative secondary feature vector based on a concentric ring approach also demonstrated poor performance on papyrus fragments. This may suggest that periodic pattern discrimination is more than simply the distance from the center of the visualized log2DFT. The position of the pixels in a log2DFT image space and the frequencies and magnitudes they represent may hold vital information which helps improve the classification.

\section{Conclusion}
The work of this article used a 2DFT based feature vector technique in the binary classification of the surface material of the DSS, achieving a performance of up to $\approx 97\%$. It offers an accurate and accessible way to classify the material of ancient historical manuscripts without the need for more labeled data (in case of neural networks) or damaging methods (for example, chemical analysis). This study provides an initial foray into using a 2DFT technique to perform material classification of such manuscripts. The ability to quickly classify the writing surface material has the potential to expedite the initial manuscript investigation. The straightforward approach presented here may be used as a starting point to help resolve any debate over the nature of a DSS fragment's material and may be applied to other ancient historical manuscripts. Furthermore, by building upon and developing the proposed system, this method demonstrates a potential for use in helping to answer more specialized questions. Examples may include intra-material classification to provide evidence for differing production techniques and/or manuscript dating. The consequences of gaining such insight by substituting in the proposed technique are threefold; the preservation of delicate ancient manuscripts from further degradation, a relatively low cost uncomplicated implementable method, and an additional extendable tool in gathering evidence to help conclude the questions surrounding the production of such manuscripts.

\section*{Acknowledgment}
The authors like to thank Mladen Popović, PI of the Hands that Wrote the Bible project, who allowed working with the data, provided valuable inputs and the labels for the materials. Finally, for the high-resolution images of the Dead Sea Scrolls, we are grateful to the Israel Antiquities Authority (IAA), courtesy of the Leon Levy DSS Digital Library; photographer: Shai Halevi.






\clearpage
\newpage
\section*{Supplement A} \label{ap:appendixA}
\begin{table}[!ht]
\caption{Plate Numbers of the Fragments used in the study.}
\centering
\begin{tabular}{lcl}
\hline \hline
\textbf{Plate number }     & \textbf{No. of Fragments} & \textbf{Material} \\ \hline \hline
\textbf{1039-2}   & 5                         & Parchment         \\ \hline
\textbf{155-1}    & 1                         & Parchment         \\ \hline
\textbf{193-1}    & 1                         & Parchment         \\ \hline
\textbf{228-1}    & 1                         & Parchment         \\ \hline
\textbf{269}      & 1                         & Parchment         \\ \hline
\textbf{489}      & 1                         & Parchment         \\ \hline
\textbf{641}      & 1                         & Parchment         \\ \hline
\textbf{974}      & 1                         & Parchment         \\ \hline
\textbf{975}      & 1                         & Parchment         \\ \hline
\textbf{976}      & 4                         & Parchment         \\ \hline
\textbf{977}      & 4                         & Parchment         \\ \hline
\textbf{978}      & 1                         & Parchment         \\ \hline
\textbf{979}      & 1                         & Parchment         \\ \hline
\textbf{5-6Hev45} & 1                         & Papyrus           \\ \hline
\textbf{641}      & 1                         & Papyrus           \\ \hline
\textbf{X100}     & 1                         & Papyrus           \\ \hline
\textbf{X106}     & 1                         & Papyrus           \\ \hline
\textbf{X130}     & 1                         & Papyrus           \\ \hline
\textbf{X207}     & 3                         & Papyrus           \\ \hline
\textbf{X304}     & 1                         & Papyrus           \\ \hline
\textbf{Yadin50}  & 1                         & Papyrus           \\ \hline
\end{tabular}
\label{tab:appendixA}
\end{table}

\section*{Supplement B} \label{ap:appendixB}
\begin{table}[!ht]
\caption{Confusion Matrix (\%) for Mean Concentric Ring Feature Vector}
\centering
\begin{tabular}{llcc}
\hline \hline
\textbf{Image type }           & \textbf{True class} & \multicolumn{2}{c}{\textbf{Classified as}} \\ \hline \hline
                               &            & Parchment        & Papyrus        \\ \hline
\multirow{2}{*}{Color}         & Parchment  & \textbf{100.0}   & 0.0            \\ \cline{2-4} 
                               & Papyrus    & 40.0             & \textbf{60.0}  \\ \hline
\multirow{2}{*}{Multispectral} & Parchment  & \textbf{87.0}    & 13.0            \\ \cline{2-4} 
                               & Papyrus    & 70.0             & \textbf{30.0 } \\ \hline
\end{tabular}
\label{tab:conf_ringmean}
\end{table}

\begin{table}[!ht]
\caption{Confusion Matrix (\%) for Standard Deviation Concentric Ring Feature Vector}
\centering
\begin{tabular}{llcc}
\hline \hline
\textbf{Image type }           & \textbf{True class} & \multicolumn{2}{c}{\textbf{Classified as}} \\ \hline \hline
                               &            & Parchment        & Papyrus        \\ \hline
\multirow{2}{*}{Color}         & Parchment  & \textbf{100.0}   & 0.0            \\ \cline{2-4} 
                               & Papyrus    & 20.0             & \textbf{50.0}  \\ \hline
\multirow{2}{*}{Multispectral} & Parchment  & \textbf{91.3}    & 8.7            \\ \cline{2-4} 
                               & Papyrus    & 50.0             & \textbf{50.0 } \\ \hline
\end{tabular}
\label{tab:conf_ringsd}
\end{table}

\begin{table}[!ht]
\caption{Confusion Matrix (\%) for the Weighted Bin Feature Vector}
\centering
\begin{tabular}{llcc}
\hline \hline
\textbf{Image type }           & \textbf{True class} & \multicolumn{2}{c}{\textbf{Classified as}} \\ \hline \hline
                               &            & Parchment        & Papyrus        \\ \hline
\multirow{2}{*}{Color}         & Parchment  & \textbf{100.0}   & 0.0            \\ \cline{2-4} 
                               & Papyrus    & 100.0             & \textbf{0.0}  \\ \hline
\multirow{2}{*}{Multispectral} & Parchment  & \textbf{95.7}    & 4.3            \\ \cline{2-4} 
                               & Papyrus    & 90.0             & \textbf{10.0 } \\ \hline
\end{tabular}
\label{tab:conf_bin}
\end{table}

\newpage
\section*{Supplement C} \label{ap:appendixC}
\begin{figure}[!ht]
\centering
\includegraphics[width=0.45\textwidth]{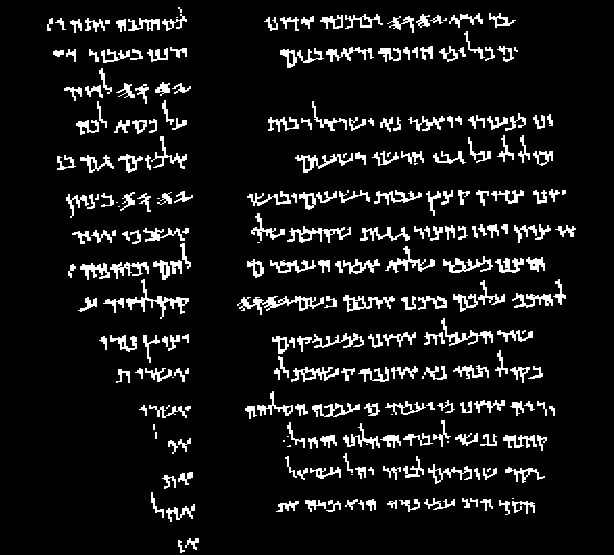}
\caption{Binary text mask taken from plate 976 in un-dilated form.}
\label{fig:apbinfragcrop}
\end{figure}

\begin{figure}[!ht]
\centering
\includegraphics[width=0.45\textwidth]{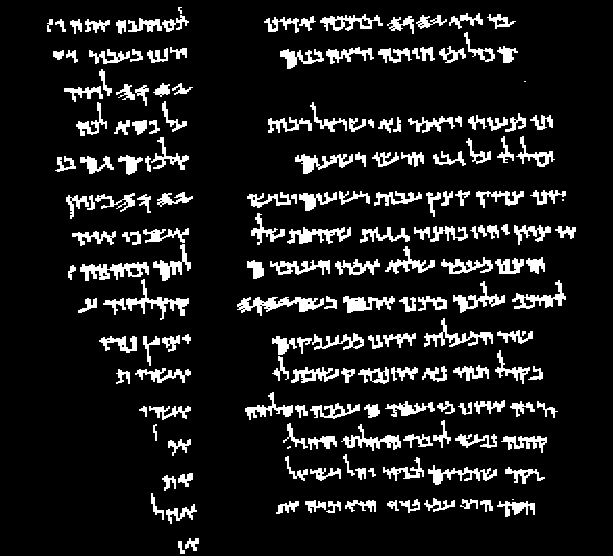}
\caption{Binary text mask taken from plate 976 after applying dilation to capture all written markings.}
\label{fig:dilate_both}
\end{figure}

\end{document}